\documentclass[conference]{IEEEtran}
\IEEEoverridecommandlockouts%Needed to identify funding in the first footnote.

\usepackage{cite}%lets us cite the bib
\usepackage{amsmath,amssymb,amsfonts}
\usepackage{algorithm}
\usepackage{algpseudocode}
\usepackage{textcomp}
\usepackage{xcolor}
\usepackage{soul}
\usepackage{graphicx} % Required for inserting images
\usepackage{comment}%lets us comment blocks of LaTeX
\usepackage{pdfpages}%lets us add pdfs
\usepackage{xurl}
\usepackage{tabularx}
\usepackage{etoolbox}
\usepackage{hyperref}
\usepackage{caption}

\newbool{Long_names}
\newbool{author_full}
\booltrue{author_full}
\newbool{longform_auth_block}
\booltrue{longform_auth_block}
\makeatletter
\newcommand{\linebreakand}{
  \end{@IEEEauthorhalign}
  \hfill\mbox{}\par
  \mbox{}\hfill\begin{@IEEEauthorhalign}
}
\makeatother

\def\BibTeX{{\rm B\kern-.05em{\sc i\kern-.025em b}\kern-.08em
    T\kern-.1667em\lower.7ex\hbox{E}\kern-.125emX}}%not sure

\title{A Survey of Transformer Enabled Time Series Synthesis\\

}
\ifbool{author_full}
{%if full auth
    \ifbool{longform_auth_block}
    {%If longform
        \author{
            \IEEEauthorblockN{
                Alexander Sommers\IEEEauthorrefmark{1}\IEEEauthorrefmark{3},
                Logan Cummins\IEEEauthorrefmark{1}, 
                Sudip Mittal\IEEEauthorrefmark{1},
                Shahram Rahimi\IEEEauthorrefmark{1},\\
                Maria Seale\IEEEauthorrefmark{2},
                Joseph Jaboure\IEEEauthorrefmark{2},
                Thomas Arnold\IEEEauthorrefmark{2}
            }
            \IEEEauthorblockA{
                \IEEEauthorrefmark{1}
                \textit{Computer Science \& Engineering} \\
                \textit{Mississippi State University}
                \\\{ams1988, nlc123\}@msstate.edu, \{mittal, rahimi\}@cse.msstate.edu
            }
            \IEEEauthorblockA{
                \IEEEauthorrefmark{2}
                \textit{Engineer Research and Development Center} \\%https://www.erdc.usace.army.mil/
                \emph{Department of Defence}
                \\\{maria.a.seale, joseph.e.jabour, thomas.l.arnold\}@erdc.dren.mil
            }
            \IEEEauthorrefmark{3}corresponding author
        }
    }
    {%if not
        \author{
        \IEEEauthorblockN{Alexander Sommers}
        \IEEEauthorblockA{\textit{Computer Science \& Engineering} \\
        \textit{Mississippi State University}\\
        ams1988@msstate.edu}
        
        \and
        
        \IEEEauthorblockN{Logan Cummins}
        \IEEEauthorblockA{\textit{Computer Science \& Engineering} \\
        \textit{Mississippi State University}\\
        nlc123@msstate.edu}
        
        \and
        
        \IEEEauthorblockN{Sudip Mittal}
        \IEEEauthorblockA{\textit{Computer Science \& Engineering} \\
        \textit{Mississippi State University}\\
        mittal@cse.msstate.edu}
        
        \linebreakand
        
        \IEEEauthorblockN{Shahram Rahimi}
        \IEEEauthorblockA{\textit{Computer Science \& Engineering} \\
        \textit{Mississippi State University}\\
        rahimi@cse.msstate.edu}
        
        \and
        
        \IEEEauthorblockN{Maria Seale}
        \IEEEauthorblockA{\textit{Eng. Research and Dev. Center} \\
        \emph{Department of Defence} \\
        maria.a.seale@erdc.dren.mil}
        
        \and
        
        \IEEEauthorblockN{Joseph Jaboure}
        \IEEEauthorblockA{\textit{Eng. Research and Dev. Center} \\
        \emph{Department of Defence} \\
        joseph.e.jabour@erdc.dren.mil}
        }
    }
}
{%if not full auth
\author{

\IEEEauthorblockN{Alexander Sommers}
\IEEEauthorblockA{\textit{Comp. Sci. \& Eng.} \\
\textit{Mississippi State University}\\
ams1988@msstate.edu}

\and

\IEEEauthorblockN{Logan Cummins}
\IEEEauthorblockA{\textit{Comp. Sci. \& Eng.} \\
\textit{Mississippi State University}\\
nlc123@msstate.edu}

\and

\IEEEauthorblockN{Sudip Mittal}
\IEEEauthorblockA{\textit{Comp. Sci. \& Eng.} \\
\textit{Mississippi State University}\\
mittal@cse.msstate.edu}

\and

\IEEEauthorblockN{Shahram Rahimi}
\IEEEauthorblockA{\textit{Comp. Sci. \& Eng.} \\
\textit{Mississippi State University}\\
rahimi@cse.msstate.edu}

\and

\IEEEauthorblockN{Maria Seale}
\IEEEauthorblockA{\textit{Engineer Research and Development Center} \\
\emph{Department of Defence} \\
maria.a.seale@erdc.dren.mil}

\and

\IEEEauthorblockN{Joseph Jaboure}
\IEEEauthorblockA{\textit{Engineer Research and Development Center} \\
\emph{Department of Defence} \\
joseph.e.jabour@erdc.dren.mil}
}
}

\begin{document}

\maketitle

\begin{abstract}
Generative AI has received much attention in the image and language domains, with the transformer neural network continuing to dominate the state of the art. Application of these models to time series generation is less explored, however, and is of great utility to machine learning, privacy preservation, and explainability research. The present survey identifies this gap at the intersection of the transformer, generative AI, and time series data, and reviews works in this sparsely populated subdomain. The reviewed works show great variety in approach, and have not yet converged on a conclusive answer to the problems the domain poses. GANs, diffusion models, state space models, and autoencoders were all encountered alongside or surrounding the transformers which originally motivated the survey. While too open a domain to offer conclusive insights, the works surveyed are quite suggestive, and several recommendations for best practice, and suggestions of valuable future work, are provided.
\end{abstract}


%intro--\/

\section{Introduction}
Generative AI methods have recently captured the public's interest with the advent of powerful multi-modal large language models, among others. Capable of generating realistic images, sound, text, and video, these models have brought about a paradigm shift in creative workflows, and much discussion of the risks and rewards such technologies pose. While applications for art, information warfare, and communication sit in the spotlight, these models also find recursive application, supporting work in AI. 

Time series data in particular are of such ubiquity and importance \cite{Large_Model_4_TS} that the dearth of generative methods in this domain, relative to others, is surprising and depriving. Time series data include the vital signs used diagnose and monitor patient conditions, signals from weather stations giving early warnings of dangerous storms, and accelerometer data from wearable devices used to detect falls. Time series models contest in the stock market, predict the survival of species, and anticipate the breakdown of machinery. With so many important aspect of life and work touched by time series data, models, and analysis, it is important that attention be paid to the application of these powerful methods to this domain. For the data-science practitioner, the value of time series generation is greater than ever. Time series generation has the potential to enable deep learning in data deprived time domain problems \cite{time_series_data_aug_survey}, preserve privacy \cite{DSAT-ECG} even while sharing data, and make more explainable, better tested, systems \cite{Based_on_CSDI}. While generative methods have existed for some time \cite{time_series_data_aug_survey}, results of their application in the ML domain suggest that generative AI are superior to the previous generation of algorithmic time series augmentations \cite{No_TGANS_and_results_of_fiddling, ACGAN_for_MFD}, and should be used if at all possible.

The transformer neural network (TNN)\cite{transformer_origin} is inextricably tied to recent advances in generative AI. Amenable to parallel training schemes, and supplied with large datasets scraped from the web, large TNN models have formed the backbone of most cutting edge generative models in the recent past. The TNN has been uncommonly adaptable, with applications in image domain and multi-modal tasks, as well as its historical target of natural language. While it has been applied to time series forecasting, less application has been made to the problem of time series generation \cite{GAI_survey}. 

A survey of the literature concerning the TNN, time series generation, and generative AI, found relatively little overlap of this network type and task. The present survey addresses this gap, providing an overview of what overlap exists, and provide lessons synthesized from that overview. While sparse compared to applications in other domains, the TNN has been introduced to the time series generative domain within and alongside other network architectures including generative adversarial networks, diffusion networks, and variational auto-encoders. Both pure and hybrid architectures exist as well. This diversity, coupled with the openness of the domain, are invitations to creative and much needed further research.

The contributions of this work are:
\begin{itemize}
    \item[$\bullet$] The identification of a gap in the survey literature at the intersection of transformers and time series generation (section \ref{sect::prior_art}).
    \item[$\bullet$] The filling of that gap, identifying twelve works (section \ref{sect::reviewed_works}) that constitute the intersection under the scope of the survey (section \ref{sect::scope}).
    \item[$\bullet$] The extraction of insights gained by viewing the disparate works in one another's context (section \ref{sect::disc}). 
    \item[$\bullet$] The provision of recommendations for best practices and future work for investigators in the domain (section \ref{sect::recom}).
\end{itemize}

The remainder of this work is organized as follows. Section \ref{sect::back} provides a brief on three background concepts needed to understand the review. Section \ref{sect::prior_art} covers prior surveys, and identifies the gap in them. It then presents the motivation to fill this gap, and the scope of the present survey. Section \ref{sect::reviewed_works} covers the works reviewed. Section \ref{sect::disc} discusses insights gained from the review, while section \ref{sect::recom} provides recommendations for investigators going forward. lastly, section \ref{sect::conc} reviews and concludes the work.

%intro--/\

%background--\/

\section{Background}\label{sect::back}

This section briefs the reader on generative neural networks, the synthesis and evaluation of time series data, and the transformer neural network.
\subsection{Generative Neural Networks}
    Artificial neural networks (ANNs) are data-driven function approximators. In image classification, the network may constitute a function $f: X \rightarrow Y$. The domain $X$ may be pictures of animals, and the codomain $Y$ class labels (e.g. cat, dog, horse). This archetype is commonly abstractive, migrating from concrete class instances to abstract evaluations (e.g. pictures of specific animals to broad class names). By contrast, generative models produce instances of concrete classes, often from abstract or meaningless inputs.

A divide in generative models concerns training supervision. 
Generative Adversarial Networks (GANs) \cite{GAN_origin} and Variational Autoencoders (VAEs) \cite{VAE_intro, VAE_tut} are mostly unsupervised, though supervised elements can be added \cite{ACGAN_origin, ACGAN_for_MFD}. GANS, unsupervised, have been particularly popular, and have been implemented using multi-layer perceptions (MLPs), convolutional neural networks (CNNs) \cite{DCGAN_origin}, recurrent neural networks (RNNs) \cite{SenseGen}, and transformer neural networks (TNNs) \cite{TransGAN, Close_competitor, Closest_competitor, Competitor_sucessor}. Diffusion models \cite{D3A-TS} and GANs can become semi-supervised when conditional generation schemes are implemented, since these conditioning signals are similar to supervisory labels \cite{cGAN_origin}. 
Conversley, the Generative Pretrained Transformer (GPT) series \cite{GPT_survey, GPT_3_origin}, and forecasting models \cite{Informer}, are trained in a supervised manner. Predicting the next element, or next several elements, in a series can easily become generative as modern Large Language Models (LLMs) demonstrate \cite{Lama}. Generation covers all domains of data, but time series offer a particular challenge.%transition sentence

\begin{figure}[ht]
\centering
\includegraphics[width=0.45\textwidth]{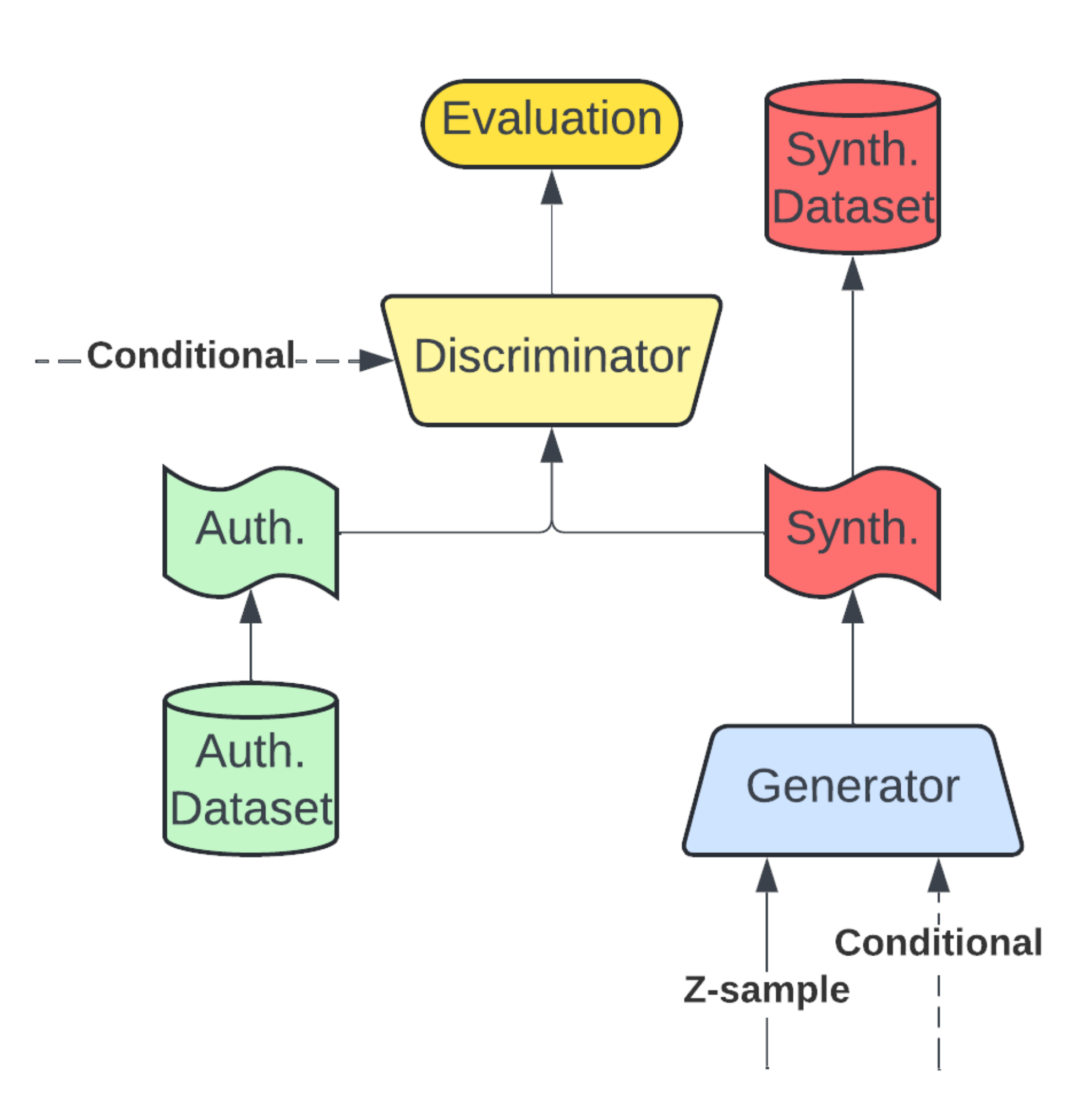}
\caption{GANs are primarily composed of two modules, the generator and discriminator. The trained generator satisfies the desiderata of the users. To apply evolutionary pressure, the discriminator is trained to distinguish synthetic and authentic samples. The adversarial game trains the generator to better deceive the discriminator, producing realistic samples. The ideal convergence scenario sees the discriminator only able to guess (50-50) as to the authenticity of a presented instance. Conditional information can be supplied at multiple points if needed.}
\label{GAN_abs}
\end{figure}
\subsection{Time Series Generation}
    A ``time series" is a set $\mathbf{S}$ of $n$ observations ordered in time s.t. $\mathbf{S} = \{s_{1}, ..., s_{n}\} \text{ and } s_{i} \in \mathbb{R}^{d}$ \cite{Time_series_def_survey}. $d$ is the dimensional of the time series with $d \ge 1$ constituting a multivariate time series. Such datasets are very common, with some of the most recognizable instances being vital signs \cite{Mimic_medical} and financial data \cite{Financial_GAN}. Evaluation of generated data is a common issue \cite{MJs_dis}, compounded in time series generation \cite{TSG_benchmark} by the variety of behavior, and high dimensionality, seen across the domain. Familiar visual subjects can be evaluated qualitatively \cite{AI_hands}, but time series are complex and foreign to our lived experiance of the world \cite{TSGAN_comparison, time_series_data_aug_survey, Temporal_latent_AE}. The two broad methods of evaluation are ``direct" and ``indirect". Direct methods of evaluation compare authentic data with its mimetic counterpart, often using statistical, or distance metrics. Indirect methods employ downstream tasks and beneficiary models trained with the synthetic data \cite{ACGAN_for_MFD, No_TGANS_and_results_of_fiddling, TSG_benchmark}. When performance of beneficiary models converges when comparing the authentic and generated data, the generated data is deemed successful.
\subsection{The Transformer Neural Network}
    The TNN is characterized by the ``attention mechanism", of which variants exist \cite{Many_attentions_1, Many_attentions_2}. Abstract understanding will suffice for this work, and will be achieved by contrasting it with corespondents in the CNN and RNN.

RNNs need to model long range dependencies as a function of the hidden state, a form of ``memory". This means the model needs to learn to remember, and to house all memories together in one representation, which is challenging. CNNs can model long range dependencies with large convolution kernels, but specificity is at odds with scale. Kernels large enough to model long range dependencies will capture many input elements. Attention mechanisms can model specific pairwise dependencies across arbitrary distances, and do so with no hidden state. The significance of the TNN to generative modeling is hard to overstate \cite{GAI_survey}, and the flexibility of the model \cite{Patching}, and scalability of its training, have made it dominant in many domains. The scalability of this training deserves a brief explanation.

\begin{figure}[!ht]
\centering
\includegraphics[width=\linewidth]{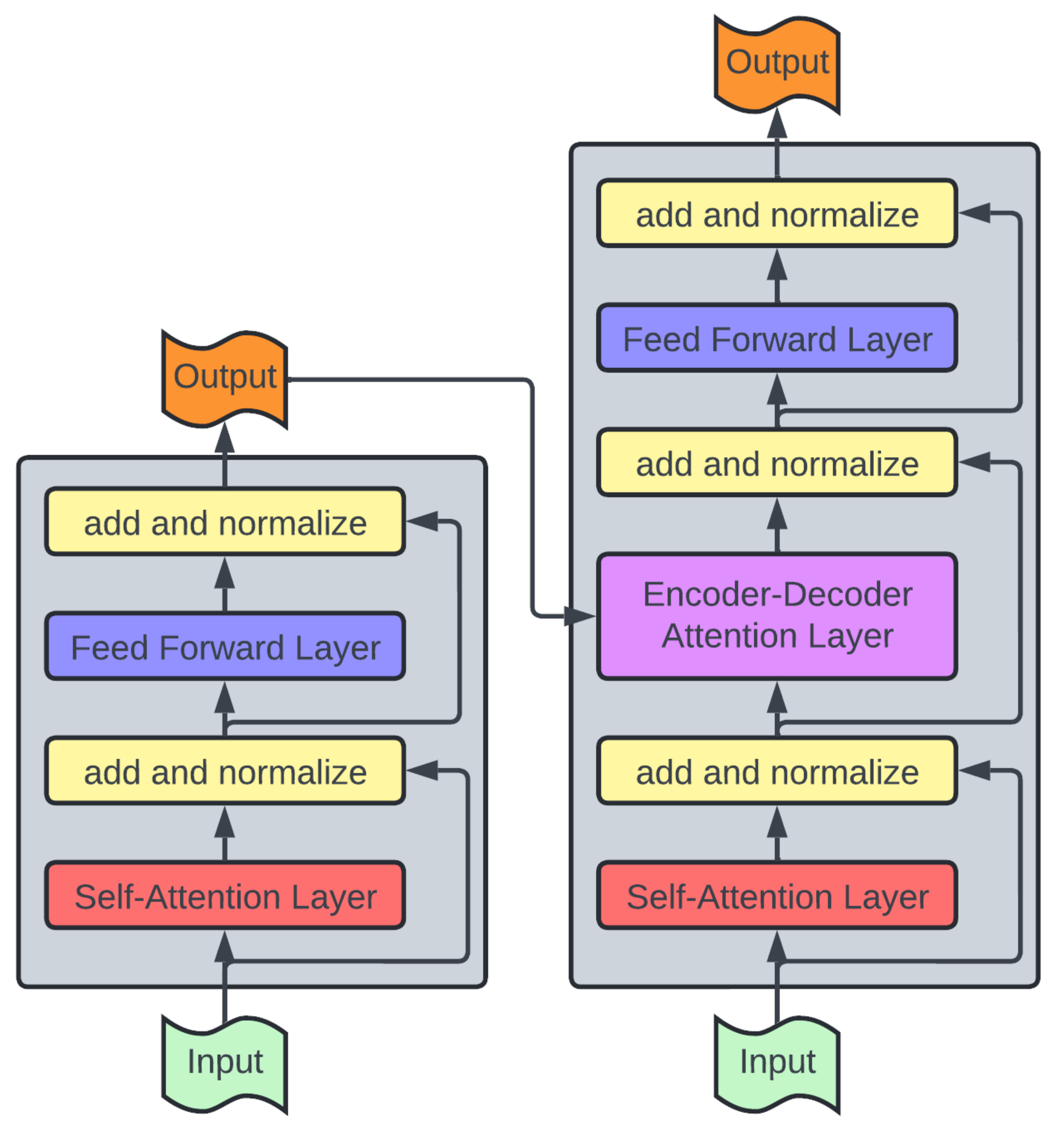}
\caption{A full TNN has two stacks, the \emph{encoder} (left) and \emph{decoder} (right), stemming from earlier sequence-to-sequence models. Both stack types are made of successive blocks, with only one block in each stack shown here. Key-Query-Value attention mechanisms project the query source as a linear combination of the Value source as a function of affinity with the Key source. Self-Attention uses the same sequence as all three sources, while Encoder-Decoder Attention uses the encoder stack output as the Key and Value source. Vaswani et al. used ``post-norm" operations, placing the normalization after each primary operation. More recent work suggests that ``pre-norm" configurations are superior \cite{postnorm_prenorm}.}
\label{TNN_abs}
\end{figure}

RNNs process sequences linearly, meaning that computing the model's output for the n\textsuperscript{th} input, the (n-1)\textsuperscript{th} output must be computed first. While this is convenient at inference time, when the network is deployed, it is inconvenient during training. Conversley, a TNN does not have this dependency, since its n\textsuperscript{th} output is a function of the 1st n elements of the input sequence, not of the n\textsuperscript{th} input and (n-1)\textsuperscript{th} output. This allows for parallel computation of outputs along a sequence, and thus parallel training. Figure \ref{TNN_abs} shows an abstract view of the foundational components of a TNN. 

The dominance of the TNN is corroborated by several recent and robust surveys, and this dominance is part of what motivates the present survey.%transition sentence.

Much prior work serves as the foundation the present survey, and section \ref{sect::prior_art} presents a summary of that work, and the gap the present survey fills.

%background--/\

%prior art--\/

\section{Prior Art and Motivation}\label{sect::prior_art}
\subsection{Prior Surveys}
The present survey was facilitated by several with much wider scopes which the reader may find valuable, especially to contextualize this work.

\cite{No_TGANS_and_results_of_fiddling}, \cite{Synthetic_data_gen_survey}, and \cite{time_series_data_aug_survey} provide an overview of time series augmentation and time series synthesis. \cite{GAN_in_TS_survey} further focuses on GANs in time series. \cite{TSG_benchmark} and \cite{TSGAN_comparison} recognize and address the challenge of comparing the wide variety of methods proposed for time series generation. \cite{Survey_of_synth_dat_gen} and \cite{GAI_survey} cover generative AI beyond time series, and provide a valuable historical and terminological background. \cite{DL_for_TSF} and \cite{TS_forecasting_survey} provide valuable reviews of deep learning time series forecasting methods. Lastly, \cite{Not_very_good_TNN_survey}, \cite{tranf_in_TS_tut}, and \cite{transf_in_TS_survey} specifically attend to TNNs in time series applications.

These surveys provide a very complete review of the constituent elements of the intersection the present survey targets. A gap identified at that intersection is articulated in section \ref{sect::gap_and_motive}.
\subsection{Motivation}\label{sect::gap_and_motive}

Recent robust surveys of the generative domain do not deeply cover time series generation \cite{GAN_survey, GAI_survey, Medical_cGANs, Synthetic_data_gen_survey, Survey_of_synth_dat_gen}. Surveys of the time domain rarely explore TNN methods \cite{GAN_in_TS_survey, TSG_benchmark, time_series_data_aug_survey, No_TGANS_and_results_of_fiddling, TSGAN_comparison, TS_forecasting_survey}. Surveys covering time series TNNs don't cover generative applications \cite{tranf_in_TS_tut, transf_in_TS_survey}. This survey seeks to fill this gap by surveying transformer enabled time series generative models. Twelve works were identified matching the standard for inclusion, filling this gap in the literature. Section \ref{sect::scope} presents that standard, which bounds the scope of this work.

Filling this gap is valuable because the deficit in time series generation coincides with the special need time domain tasks have for it \cite{TSG_benchmark, TS_forecasting_survey}. Iglesias et al. \cite{time_series_data_aug_survey} argue that deep learning practitioners \cite{Depth_improves_accuracy, DL_for_TSF} would be major beneficiaries of better generation methods because deep models require substantially more data to train \cite{LFD_complexity_cost}. 

\begin{figure*}[ht]
\centering
\includegraphics[width=1\textwidth]{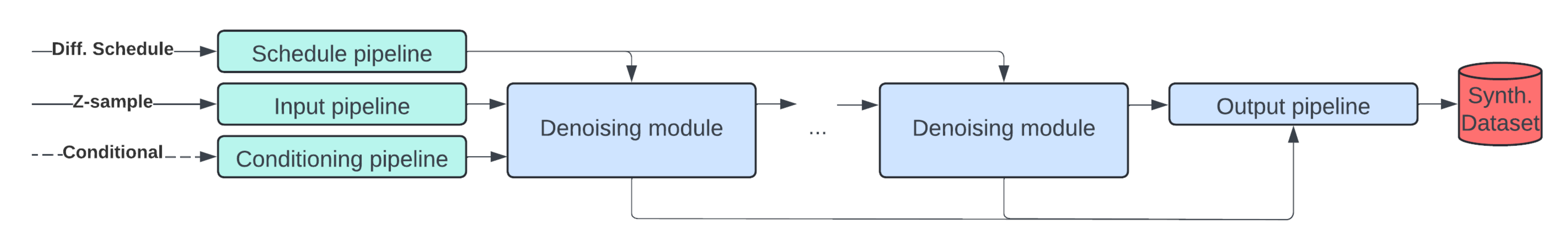}
\caption{Diffusion models in the DiffWave lineage typically consist of several denoising blocks, each with a skip connection. When generation is conditioned, conditional information is supplied alongside the noise input. Noise is drawn from the standard normal distribution $\mathbf{Z}$. An embedding indicating each step in a diffusion schedule is supplied to each module, while only the first module accepts the noise input and any conditional data. Each input has its own embedding pipeline. All skip connection outputs are accepted, as is the final denoised output, by the final output module.}
\label{Diffusion_abs}
\end{figure*}

Synthetic data can also enable counterfactural exploration \cite{Temporal_latent_AE}, explainability \cite{Logan_PM_XAI_survey}, and allow derivatives of private data to be shared \cite{SenseGen, DSAT-ECG}, especially important in medical settings. The recent dominance of the TNN on sequence tasks prompts the present survey.
\subsection{Scope}\label{sect::scope}
The scope of this survey is set by a definition of the narrow intersection of time series generation and TNN enabled models. This section provides these definitions, after which section \ref{sect::reviewed_works} reviews selected works.
\subsubsection{Definition of Generation}
Three tasks cover generation in time series: imputation, forecasting, and synthesis; each will be briefly described. 

Imputation, filling in missing values, faces inward and so is not focused on here. Forecasting, predicting future elements, is quite common, and can be repurposed for generation, especially in an autoregressive manner. ``Synthesis", used interchangeably with ``generation" \cite{WGAN_2, Competitor_sucessor, TSG_benchmark}, is the fabrication of data without reference to a prior ground truth. The primary targets of the survey were generative methods, but some forecasting methods are included because of their leveraging of generative methods, or the precedent set by LLMs of forecasters being turned into synthesizers \cite{Lama}.
\subsubsection{Definition of Transformer Enabled}
Transformer enabled neural models are either dominated by the TNN, or have it play a significant role. Again, this focus is due to the elsewhere dominance of the TNN, but the apparent under-utilization in time series generation.

%prior art--/\

%core review--\/

\section{Transformer Enabled Time Series Generative Models}\label{sect::reviewed_works}
This section provides brief descriptions of the works constituting the target subdomain. Readers are encouraged to refer to the works themselves for greater detail. After these works are reviewed, section \ref{sect::disc} provides insights gained from viewing them in context. Table \ref{data_tab} lists the datasets used, and Table \ref{tax_tab} provides a taxonomy of the methods surveyed.

\begin{table}[h]
\centering
\begin{tabularx}{.85\linewidth}{|l|l|p{.25\linewidth}|}
\hline
\textbf{Method} & \textbf{Task Type} & \textbf{Architecture Type} \\ \hline

Informer* & Forecasting & TNN dominated \\ \hline

AST* & Forecasting & TNN dominated \\ \hline

ACT* & Forecasting & Hybrid \\ \hline

GenF* & Forecasting & Hybrid \\ \hline

TTS-GAN & Synthesis & TNN dominated \\ \hline

TsT-GAN* & Synthesis & TNN dominated \\ \hline

TTS-CGAN & Synthesis & TNN dominated \\ \hline

MTS-CGAN & Synthesis & TNN dominated \\ \hline

Time-LLM & Forecasting & TNN dominated \\ \hline

DSAT-ECG & Synthesis & Hybrid \\ \hline

Time-Transformer AAE & Synthesis & Hybrid \\ \hline

TimesFM & Forecasting & TNN dominated \\ \hline

Time Weaver & Synthesis & Hybrid \\ \hline
\end{tabularx}
\caption{A practice taxonomy of surveyed models concerns principally their target task, and the dominance of TNN modules within the architecture.\\\textbf{\small{*Model impliments countermeasures to auto-regressive error accumulation}}}
\label{tax_tab}
\end{table}

\subsubsection{Informer}
Zhou et al.'s Informer \cite{Informer} is a full encoder-decoder TNN forecaster. The authors develop probabilistic sparse attention (PSA) to leverage sparsity in the affinity matrix, avoiding computing dot products on queries with homogeneous affinity across keys. The encoder accepts historical data, and the decoder accepts a truncated history padded with a placeholder. The output is the truncated history concatenated with the forecast, achieving multi-horizon forecasting. This ``generative style" forecast avoids accumulating autoregressive error and made the model subject to the survey. Informer is one of the few methods surveyed that other methods were compared to: TimesFM, Time-LLM, and GenF all compare their performance to that of Informer, highlighting its significance.
\ifbool{Long_names}
{
\subsubsection{Adversarial Sparse Transformer (AST)}
}
{
\subsubsection{AST}
}
Wu et al.'s \cite{AST} AST is a hybrid multi-time-horizon forecaster with an auxiliary discriminator. Its encoder-decoder TNN uses entmax sparse attention \cite{ent_max} and acts as a generator competing with the discriminator. Supervised forecasting training sees the encoder accept historical time series and ``covariates". These are separate, but temporally concurrent data which may be static or vary dynamically. The decoder accepts covariates corresponding to the target time horizons. Adversarial unsupervised training regularizes predictions by balancing plausible forecasts from of the historical data with plausibility of the predicted time series as members of the dataset. This is avoid the accumulation of error seen in next-step ``autoregressive" forecasters. Its combination of the generative and forecasting pressures made it relevant to the present work. Its inclusion of covariates places it closer to tabular data, and also makes it similar to conditioned generative models. The ACT \cite{ACT} model follows AST, adding informer's use of placeholders and the ``convolutional attention block", combining a TCN and an encoder transformer module.
\ifbool{Long_names}
{
\subsection{Generative Forecasting (GenF)}
}
{
\subsubsection{GenF}
}
Liu et al.'s GenF \cite{TGAN_uses_informer} is a hybrid generative forecasting model. Addressing error accumulation in autoregressive forecasting, the authors mathematically devise a compromise between autoregressive and direct prediction. When predicting a time horizon $h | h \geq 1$, some horizon $l|l\leq h$ is selected and $l$ steps are generated autoregressively. Then the target horizon $h$, only $h-l$ steps from the now synthetic ``end of history", is directly predicted. The canonical TNN encoder is used as the direct forecasting method, and the framework can accommodate other forecasters.
\ifbool{Long_names}
{
\subsubsection{Transformer-based Time-Series Generative Adversarial Network (TTS-GAN)}
}
{
\subsubsection{TTS-GAN}
}
Li et al.'s TTS-GAN \cite{Close_competitor} is as pure TNN GAN \cite{GAN_origin}. Building on \cite{TransGAN} and \cite{Patching}, the authors target the synthesis of bio-signal data, which is often hard to get. This model founded one of the main lineages in this subdomain, with two other methods directly building on it, and several citing it.
\ifbool{Long_names}
{
\subsubsection{Time-series Transformer Generative Adversarial Network (TsT-GAN)}
}
{
\subsubsection{TsT-GAN}
}
\begin{figure*}[h]
\centering
\includegraphics[width=1\textwidth]{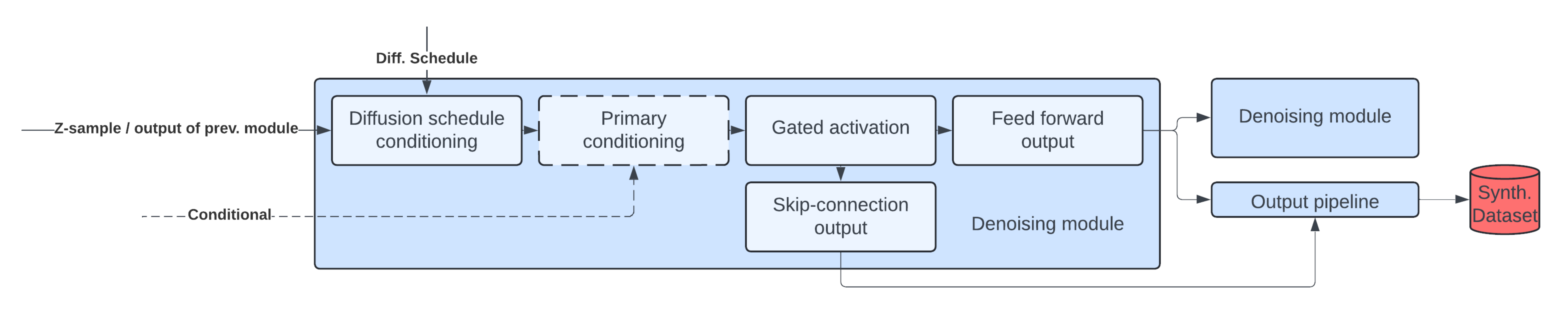}
\caption{This is an abstract view of a denoising block as used in diffusion models of the DiffWave lineage. A view of this module in situ with its neighbors can be seen in figure \ref{Diffusion_abs} . The exact implementation  of the submodules (CNNs, TCNs, TNNs, SSMs, etc...) can vary, so here they are represented as ``processes". The first process conditions the input to the block with the time step in the noise/diffusion schedule (\cite{DiffWave} provides more details on diffusion schedules, and the ``forward" and ``reverse" processes). If conditioned generation is implemented, and this is the first block, then a primary conditioning process is supplied. The gated activation process, inherited from WaveNet \cite{WaveNet} feeds both outputs. The feed forward process goes to the next block, or the output, and the skip connection jumps directly to the output pipeline, which accepts all skip connections.}
\label{Denoise_abs}
\end{figure*}

Srinivasan et al.'s TsT-GAN \cite{TsT-GAN} is a pure TNN GAN. The authors attempt to balance the modeling of next-step dependencies with global pan-sequence dependencies using multi step training and a four part model (generator, embedder, predictor, and discriminator). Building on \cite{TimeGAN}, the embeddor and predictor are trained in a next-step prediction task. The generator is then trained on a BERT style masked-position imputation task\cite{BERT_origin}, and then on a generation task where it must use the predictor layer as an intermediate between it and the discriminator. This combines the adaptive pressures of generating globally plausible sequences and plausible autoregressive transitions within those sequences. The results are evaluated using downstream tasks.
\ifbool{Long_names}
{
\subsubsection{Transformer-based Time-Series Conditional Generative Adversarial Network (TTS-CGAN)}
}
{
\subsubsection{TTS-CGAN}
}
Li et al.'s TTS-CGAN \cite{Closest_competitor}, an extension of TTS-GAN, is a pure TNN conditional-GAN (cGAN) \cite{cGAN_origin}. They use the asymmetrical conditioning scheme proposed in \cite{ACGAN_origin}, used in a CNN based time series application in \cite{ACGAN_for_MFD}. The Wasserstein-loss scheme \cite{WGAN_origin}, a gradient penalty version \cite{WGAN_2}, is also employed. The core method of evaluation is a derivative of Wavelet Coherence (WTC) \cite{Wavelet_coher}.
\ifbool{Long_names}
{
\subsubsection{Multivariate Time Series Conditional Generative Adversarial Network (MTS-CGAN)}
}
{
\subsubsection{MTS-CGAN}
}
Mandane et al.'s MTS-CGAN \cite{Competitor_sucessor} builds on TTS-GAN. It uses standard cGAN conditioning \cite{cGAN_origin}, but can accept both class labels and time series projections as conditionals. The hyperparameter $\alpha$ is introduced, to weigh between the influence of the conditioning information and the sample of the latent space. They also develop a time series adaptation of the Fréchet inception distance to evaluate output quality and achieve early stopping.
\ifbool{Long_names}
{
\subsubsection{Time-Large Language Model(Time-LLM)}
}
{
\subsubsection{Time-LLM}
}
Jin et al.'s Time-LLM \cite{Time_lama} applies the capabilities of pretrained LLMs to forecasting. Their innovation is the application of cross modal (text-time series) transduction modules to ``reprogram" a frozen pretrained LLM (Llama-7B by default). The first of the two modules is used to embed the time series data using ``text prototypes", derivatives of word embeddings, and also accepts a ``prompt-as-prefix" to contextualize the forecasting task. The second module accepts the output of the LLM and collapses it into a number for next-step forecasting. The success of the method suggests the possibility that the few-shot capabilities such models possess might be leveraged into non-language domains.
\ifbool{Long_names}
{
\subsubsection{Diffusion-Based State Space Augmented Transformer-Electrocardiogam (DSAT-ECG)}
}
{
\subsubsection{DSAT-ECG}
}
Zama et al.'s DSAT-ECG \cite{DSAT-ECG} is a conditional generator employing diffusion and state space models (SSMs). It belongs to the WaveNet lineage, backwards through SSSD\textsuperscript{S4} \cite{SSSD_S4}, DiffWave \cite{DiffWave}, and WaveNet \cite{WaveNet}. It replaces the S4 blocks of \cite{SSSD_ECG} with SPADE \cite{SPADE} blocks, which contain a frozen SSM block and several TNN encoders with local attention variants. SPADE blocks are a hybrid architecture which employ TNNs to model local dependencies and the SSM to model global dependencies.
\ifbool{Long_names}
{
\subsubsection{Time-Transformer Adversarial Autoencoder (Time-Transformer AAE)}
}
{
\subsubsection{Time-Transformer AAE}
}
Liu et al.'s Time-Transformer Adversarial Auto Encoder (AAE) \cite{TNN_TS_AE} is a synthesizer using the autoencoder archetype. The AAE \cite{AAE_origin} is an AE where the encoder module doubles as a generator contending with an auxiliary discriminator. As with variational autoencoders \cite{VAE_origin}, AAEs map from the domain to the set of Gaussian distributions. Both methods balance reconstruction loss with the adherence of mappings to the standard normal distribution, $\mathcal{Z}$. The authors introduce the Time-Transformer module, composed of parallel stacks of temporal convolutional networks (TCNs) and TNN encoders that are linked in every block by bidirectional cross attention \cite{Cross_attention_origin}.

\ifbool{Long_names}
{
\subsubsection{Time Series Foundation Model(TimesFM)}
}
{
\subsubsection{TimesFM}
}
Das et al.'s TimesFM \cite{TimesFM} is a decoder only time series foundation model for forecasting. The phrase ``decoder only" comes from the original encoder-decoder architectures. There, the later stack produced the output. Decoder only models lack encoder-decoder attention, and instead have an embedding module prior to the stack of encoder-style TNN blocks. TimesFM predicts patches of outputs (multi-horizon forecasting). To achieve a dataset large and representative enough to pretrain the foundation model, the authors collected roughly one billion data points of Google search trend data, one hundred billion data points of Wikipedia page-view data, and several others. They also used artificial data, such as sinusoidals, with details given as to the process. Foundation models form the core of many generative methods, especially cross modal ones, and so TimesFM was included in this survey.
\subsubsection{Time Weaver}
Narasimhan et al.'s Time Weaver \cite{TimeWeaver} is an extension of the CSDI and SSSD diffusion models. They contribute a conditioning pipeline accepting of heterogenious ``metadata", time varying categorical and continuous variables concurrent with the time series data. These are converted into a conditioning vector via a series of embedding and self-attention layers, trained using contrast learning \cite{Contrast_learning_origin}. The metadata are similar to the ``covariates" in AST. The CSDI backbone contains two TNN encoders which model inter-feature, and inter-temporal, dependencies respectively. The authors propose the Joint Frechet Time Series Distance (J-FTSD) to evaluate the performance of conditioned time series generation models.
\subsubsection{Honorable mention}
The narrow focus of the survey, and other constraints, require that many broadly relevant works be excluded, of which some must at least be mentioned.
In \cite{TFT}, Lim et al. present the Temporal Fusion Transformer. This forecasting model proposed an innovative combination of attention mechanisms, gated residual networks, and LSTMs. A highly hybridized network, it showcases the capability such models can have, and its performance highlights the possibility of systems where components are chosen to carefully align with distinct responsibilities. Further, it and PSA-GAN \cite{PSA-GAN} highlight the value of self attention as a discrete module which can be used apart from the remainder of standard transformer components.

In \cite{CSDI}, Tashiro et al. present CSDI, an imputation and forecasting model, one of several diffusion based models in the DiffWave/WaveNet Lineage. It includes two encoder style transformer layers that are used to model intra-feature temporal dependencies and inter-feature dependencies seperatley. This is to say that one transformer layer treats the concurrent features in a time series as separate, and models temporal dependencies within each isolated signal. The other transformer models dependencies between features while treating each time step as separate.  This would go on to merit inclusion in Time Weaver.

In \cite{TimeGAN}, Yoon et al. propose TimeGAN, providing a valuable baseline to which later works were compared, and innovating with its jointly trained latent embedding space. This approach, and others, attempt to deal with the challenge of modeling multivariate time series by finding a more compact representation of the features at each time step \cite{Temporal_latent_AE}. Its longevity as a baseline has been the most as close to a benchmark as the subdomain had until the work of Ang et al. \cite{TSG_benchmark}.

Lastly, in \cite{Based_on_CSDI}, Coletta et al. propose DiffTime, demonstrating the possibility of quite specific conditional generation. Diffusion models may come to dominate conditional generation tasks, having already shown promise in the image domain \cite{GAI_survey}.

\begin{table}
\centering
\caption{Dataset usage across included works}
\begin{tabular}{|c c|}%Table of datasets used in training and/or testing a model
    \hline
    Dataset & Works\\
    \hline
    Synth. by authors & \cite{TNN_TS_AE, TimesFM, sEMG, TsT-GAN}\\
    \hline
    Electrical load \cite{ECL} & \cite{Informer, Time_lama, TimesFM, TimeWeaver}\\
    \hline
    Transformer oil temperature \cite{Informer} & \cite{Informer, Time_lama, TimesFM}\\
    \hline
    Smartphone accelerometers \cite{Smart_accel_dat} & \cite{Close_competitor, Closest_competitor, Competitor_sucessor}\\
    \hline
    Air quality \cite{beijing_air} & \cite{TGAN_uses_informer, TsT-GAN, TimeWeaver}\\
    \hline
    NOAA climate  \cite{NOAA_climate} & \cite{Informer, TimesFM}\\
    \hline
    M4-competition-2018 \cite{M4_comp} & \cite{Time_lama, TimesFM}\\
    \hline
    Physionet \cite{Physionet} & \cite{Close_competitor, Closest_competitor}\\
    \hline
    Single Household electricity use \cite{UCI_appliances_energy, UCI_appliances_energy_origin} & \cite{TNN_TS_AE, TsT-GAN}\\
    \hline
    Alphabet Inc. stock prices \cite{ggl_stock_dat, TimeGAN} & \cite{TNN_TS_AE, TsT-GAN}\\
    \hline
    Electrocardiography \cite{ECG} & \cite{DSAT-ECG, TimeWeaver}\\
    \hline
    MIMIC medical \cite{Mimic_medical} & \cite{TGAN_uses_informer}\\
    \hline
    Fossil fuel consumption \cite{World_Bank_fuel_data} & \cite{TGAN_uses_informer}\\
    \hline
    Multi-household electricity use \cite{German_household_electricity} & \cite{TGAN_uses_informer}\\
    \hline
    MPI Biochem. weather  \cite{MPIB_climate} & \cite{Time_lama}\\
    \hline
    California Traffic \cite{CA_dot_traffic_data} & \cite{Time_lama}\\
    \hline
    Minneapolis Traffic \cite{Minneapolis_traffic_data} & \cite{TimeWeaver}\\
    \hline
    Influenza like illnesses\cite{CDC_influenza} & \cite{Time_lama}\\
    \hline
    M3-competition-2000 \cite{M3_comp} & \cite{Time_lama}\\
    \hline
    Google search trends \cite{ggl_srch_trnd} & \cite{TimesFM}\\
    \hline
    Wikipedia page-views \cite{Wiki_pg_view} & \cite{TimesFM}\\
    \hline
    LibCity dataset collection \cite{LibCity} & \cite{TimesFM}\\
    \hline
    Monash dataset collection \cite{Monash} & \cite{TimesFM}\\
    \hline
    Darts dataset collection \cite{Darts} & \cite{TimesFM}\\
    \hline
    MIT-BIH arrythmias \cite{MIT_arryth} & \cite{Closest_competitor}\\
    \hline
    Pachelbel's Canon (Canon in D) & \cite{TNN_TS_AE}\\
    \hline
    Electrocochleography \cite{ECochG} & \cite{TNN_TS_AE}\\
    \hline
    synth. Electromyography \cite{sEMG, EMG_closedform} & \cite{sEMG}\\
    \hline
    Spatiotemporal Chickenpox cases \cite{Spatio_temp_CP} & \cite{TsT-GAN}\\
    \hline
\end{tabular}
\label{data_tab}
\end{table}

%core review--/\

%insights, discussion, suggestions/problems/future work--\/

\begin{figure}[h]
\centering
\includegraphics[width=.5\textwidth]{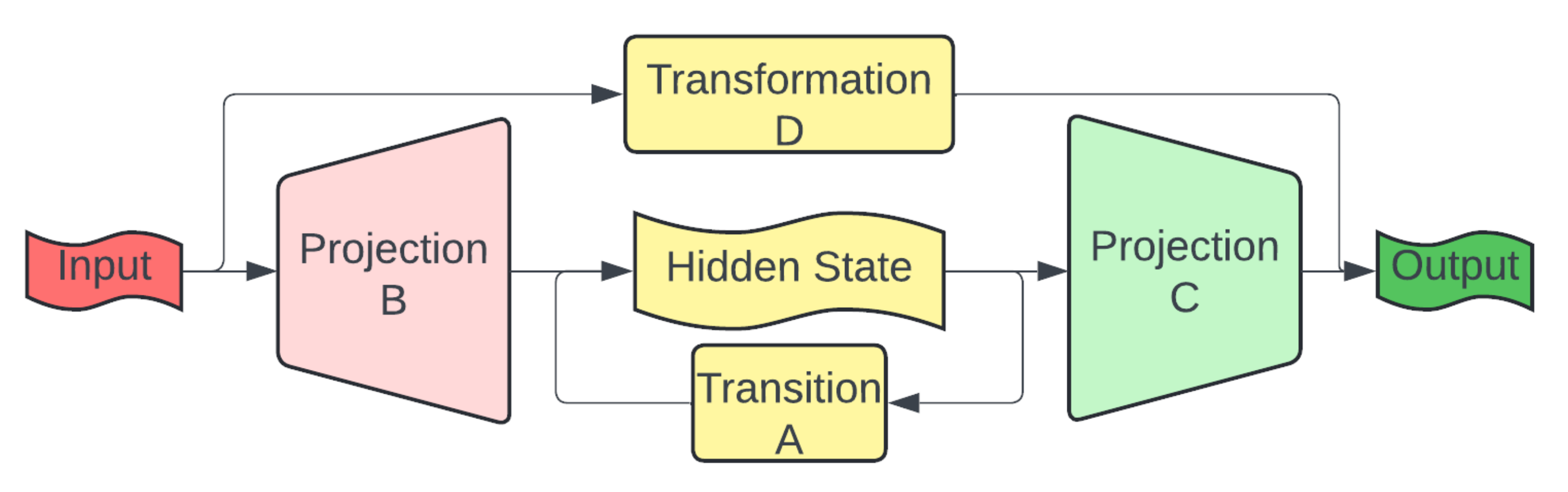}
\caption{A single state space module executes a three part process of projection, evolution, and a skip connection. Projection matrix B casts a sequence to a higher dimensionality. This projection is treated as coordinates in a state space, a vector space describing the attributes of a dynamic system. The transition matrix A applies several partial differential equations to this state vector, progressing it forward in time by a discrete time step. The C projection returns the evolved state to the original dimensionality. Transformation D acts as a learnable skip connection. With trained parameters, this facilitates an analog to an RNN's autoregressive predictions, but can be trained like a CNN, with a great deal of precomputation, as achieved in \cite{S4}.}
\label{SSM_abs}
\end{figure}

\section{Discussion}\label{sect::disc}
Several conclusions can be drawn on the basis of the reviewed material. Several are profitably actionable for the interested researcher, and suggest future lines of investigation that could benefit the domain, and those which would employ its products.

First, and over all, time series generation appears quite underinvestigated relative to its peer subdomains. There is also a sense of decoherence in the domain. While several works surveyed referenced similar backgrounds, and some of the more recent even referenced a few of the earlier works, most works seem only aware of their most immediate prior. There was also occasional duplication of work, variance in notation, and the repeated creation of evaluation metrics. The variety of approaches, though, suggests opportunity, as practice has not yet converged on a single most respected method.

That said, two lineages seem apparent in the reviewed works. Firstly, GANs, building here on TTS-GAN and/or TimeGAN, are familiar to many researchers. Though challenging to train, this archatype of model is now one of the oldest, best understood, and most flexiblly extended in the generative domain. Figure \ref{GAN_abs} depicts an archetypal GAN. Much progress has been made using GANs, especially in the image domain, where things like style transfer \cite{Style_transfer} were first developed. Secondly, diffusion models, descended here from DiffWave, are a more recent innovation. They suffer fewer training instabilities, but are novel relative to GANs, and the generative community is less familiar with them. Figures \ref{Diffusion_abs} and \ref{Denoise_abs} depict a diffusion model and denoising module respectively. While slower at inference time than GANs, many applications of generative models are not highly time sensitive. The training methodology for these networks is somewhat simpler compared to GANs as there is only one network actually at play. A third category may or may not develop its own distinct body from diffusion models, ``state-space models". SSMs are an old sequence modeling method with a long history in physics and control theory. They have an emerging relevance in sequence modeling work as a result of recent innovations in their learning capacity and computational efficiency \cite{S4, Mamba_origin}. This survey encountered SSMs in and outside of diffusion models, and the extent of their capabilities is an area of ongoing research. Figure \ref{SSM_abs} gives an abstract depiction of an SSM.

Deeper insight into best practices, and greater rigor in evaluating the available methods, await the development and adoption of a shared performance benchmark. The absence of this standard has likely contributed to the accidental reinvention of certain conditioning schemes \cite{TimeWeaver, Close_competitor}, and the invention of at least three novel performance metrics \cite{Close_competitor, TimeWeaver, Competitor_sucessor}. While only the informer forecaster was successfully compared directly to other reviewed methods, a comparison of Time Weaver and TTS-GAN was attempted, with the latter failing to converge \cite{TimeWeaver}. Ang et al.'s benchmark \cite{TSG_benchmark} would be a excellent standard for works in the domain to adopt, provided it could be updated to accommodate conditional generation \cite{TimeWeaver}. Conditional generation is so commonly practiced that a benchmark without it may be unlikely to see wide adoption.

Given that rigorous comparison awaits benchmarking, only preliminary recomendations can be offered, and are found in section \ref{sect::recom}. 

There is great opportunity and need for future work in the application of TNNs to time series generation, and also in time series generation as a whole. Not only are these domains understudied, but the variety of approaches seen suggest an unsettled question. While TNNs have been the last word in NLP work for some time (only recently challenged by SSMs), there is not a clear dominant archetype yet, and the two lineages this survey discovered might not tell the whole story. Specific suggestions for future work are provided in section \ref{sect::recom}.

\section{Recommendations}\label{sect::recom}
This section provides recommendations both as to specific practices and future lines of inquiry. Investigators are encouraged to refer to the works reviewed for implementation details.

As stated in the previous section, rigorous comparison awaits a well developed and adopted benchmark. In fact, the execution of several methods across such a benchmark, without any additional invention on behalf of the investigators involved, would be very valuable to the field. \cite{No_TGANS_and_results_of_fiddling} provides an excellent standard of rigor to which any such work might aspire, and would be compatible with the benchmark proposed in \cite{TSG_benchmark} which adopts the common ``train synthetic, test real" paradigm of indirect evaluation.

Even lacking this standard of comparison, the reviewd works suggest some bestpractices. Hybrid diffusion architectures seem superior to GANs in simplicity of design and stability of training. While the GAN-archetype of generation is sub-module agnostic (it can be implemented with CNNs, RNNs, and TNNs, for example), the hybrid nature of these successful models is worth noting. It is true that extensivley trained, relativley pure TTN archatectures have reigned supreme for some time now. To quote Dosovitskiy et al. \cite{Patching} ``\emph{large scale training trumps inductive bias}". However, the witnessed hybrid models suggest the continued utility of inductive bias, especially for task specific models. Pairing sub-architectures with sub-tasks seemed to have benefited all methods that employed it. This may be especially relevant in a domain where data relevant to a use case may be less abundant than in the NLP or image domains. Large scale training can only trump inductive bias if available data makes it an option. Leveraging inductive bias is by no means incompatible with large scale training either, and in use-cases such as finance, where small fractions of improvement can differentiate competitors, any edge is welcome.

After the top priority of a shared benchmark, a near second is exploring and evaluating the transferability of generative models from large to small datasets. This is especially relevant to data augmentation use cases. An ideal scenario, and thus one worthy of investigation, sees generative models pretrained on large time series datasets, and then being tuned on smaller ones for a particular generative use case. This would be very much analogous to the way pretrained LLMs are used today in a huge variety of NLP applications.

Of secondary, but related interest is a time series analog of style transfer \cite{Style_transfer}, which could be very valuable in counterfactual and explainability work, and also in creating hypothetical data used to test other systems, such as medical of machine malfunction diagnosis.

Continued exploration of how to represent time series data in these models would also be valuable, as would comparison of already used methods, mostly involving patching. TNNs were originally introduced in language domain tasks where word embeddings \cite{GloVe} are used to make tokens ``meaningful" to the network. Natural language is semantically dense, with each token possessing significant meaning, even before considering how that meaning shifts in the context of the surrounding tokens. Time series data seem not to possess this density, and so patching \cite{Patching} and reprogramming \cite{Time_lama} are employed. These methods map several time steps, semantically sparse on their own, to a more semantically dense gestalt. The clear boost to language domain tasks that word embedding had when first introduced suggests that this insight, and evolutions on it, will be part of increased performance in time series synthesis, and other sequence modeling tasks.

 lastly, the variety of architectures, and scales of window/horizon, seen in the reviewed works asks an important question. Which architecture is most suited for modeling dependencies over which temporal range? Though requiring much more research, this survey suggests that TCNs may be best for short ranges (tens of time steps), and SSMs for long range (thousands of time steps), with flexible attention mechanisms tuned to middle ranges (hundreds of time steps). If this is born out, cross-attention \cite{Cross_attention_origin} mechanisms may be useful for synchronizing these short, medium, and long range views of time with one another.

\section{Conclusion}\label{sect::conc}
This survey has explored the valuable and sparsely populated intersection of the TNN and time series generation. The ubiquity of time series data, the utility of high quality conditional synthesis, and the openness of not just TNN application, but the whole of the time series generation domain, all promise great reward to interested investigators. Flanked on all sides by surveys of much larger domains, the subject of the present survey none the less suggests a number of unanswered questions, and supports the continued relevance of hybrid models and inductive bias, and the emerging relevance of SSMs. The variety of models on display also invites investigators of a variety of specialties to collaborate, with the best practices of the domain still very much up for debate. As the domain advances, time series generation will take its place alongside image and language domain generation in augmenting the creative capacity of our work-forces, and will offer valuable tools for testing and explaining our systems. The opportunities this presents to the investigator and user alike are very inviting.

%insights, discussion, suggestions/problems/future work--/\

%acknowledgement--\/

\section{Acknowledgement}
This material is based upon work supported by the Engineering Research and Development Center - Information Technology Laboratory (ERDC-ITL) under Contract No. W912HZ23C0013. Any opinions, findings and conclusions or recommendations expressed in this material are those of the author(s) and do not necessarily reflect the views of the ERDC-ITL.

%acknowledgement--/\

%bib settings--\/

\bibliographystyle{ieeetr} % We choose the "plain" reference style
\bibliography{ref} % Entries are in the refs.bib file

%bib settings--/\

\end{document}